\documentclass[dvipsnames]{article} %
\usepackage{colm2024_conference}

\usepackage{booktabs}
\usepackage{graphicx}
\usepackage{enumitem}
\usepackage{wrapfig}
\usepackage{algorithm}
\usepackage{algpseudocode}

\usepackage{microtype}
\usepackage{amsmath}
\usepackage{colortbl}
\usepackage[utf8]{inputenc}
\definecolor{lightgray}{rgb}{0.9,0.9,0.9}
\usepackage{caption}
\usepackage{subcaption}
\usepackage{setspace}
\usepackage{url}
\usepackage{multirow}
\usepackage{colortbl}
\usepackage{tabularx}
\usepackage{blindtext}
\usepackage{pgfplots}
\pgfplotsset{compat=1.18} 
\usepackage{tikz}
\usetikzlibrary{er,positioning,bayesnet}
\usepackage{makecell}
\usepackage{tipa}
\usepackage{siunitx}
\usepackage{nicefrac}
\usepackage{tocloft}
\usepackage{listings}
\usepackage[raster,skins]{tcolorbox} %
\usepackage{xltabular}
\usepackage{adjustbox}
\usepackage{xurl}
\usepackage{rotating}
\usepackage[normalem]{ulem}
\usepackage{multicol}
\usepackage{titlesec}
\usepackage{helvet}
\useunder{\uline}{\ul}{}

\usepackage{amsmath,amsfonts,bm}

\def\eqref#1{equation~\ref{#1}}

\def\1{\bm{1}}

\DeclareMathAlphabet{\mathsfit}{\encodingdefault}{\sfdefault}{m}{sl}
\SetMathAlphabet{\mathsfit}{bold}{\encodingdefault}{\sfdefault}{bx}{n}

\renewcommand{\ghlink}{https://github.com/alibaba/ROLL}

\newcommand*\justify{%
  \fontdimen2\font=0.4em
  \fontdimen3\font=0.2em
  \fontdimen4\font=0.1em
  \fontdimen7\font=0.1em
  \hyphenchar\font=`\-
}

\renewcommand{\texttt}[1]{%
  \begingroup
  \ttfamily
  \begingroup\lccode`~=`/\lowercase{\endgroup\def~}{/\discretionary{}{}{}}%
  \begingroup\lccode`~=`[\lowercase{\endgroup\def~}{[\discretionary{}{}{}}%
  \begingroup\lccode`~=`.\lowercase{\endgroup\def~}{.\discretionary{}{}{}}%
  \catcode`/=\active\catcode`[=\active\catcode`.=\active
  \justify\scantokens{#1\noexpand}%
  \endgroup
}

\newcommand{\SysName}{\texttt{ROLL}}

\newenvironment{denseitemize}{
	\begin{itemize}[topsep=2pt, partopsep=0pt, leftmargin=1.5em,label=$\bullet$]
		\setlength{\itemsep}{3pt}
		\setlength{\parskip}{0pt}
		\setlength{\parsep}{0pt}
	}{\end{itemize}}

\title{Reinforcement Learning Optimization for Large-Scale Learning: \\ An Efficient and User-Friendly Scaling Library}

\author{
\bf ROLL Team
}

\begin{document}

\maketitle

\begin{abstract}
The remarkable success of Reinforcement Learning (RL) in advancing Large Language Models (LLMs) has spurred the development of efficient RL training frameworks. These frameworks, however, entail the coordinated management of multiple models and multi-stage pipelines, presenting challenges in efficiency, scalability, and usability.
To respond, we introduce \SysName{}, an efficient, scalable, and user-friendly library designed for \textbf{R}einforcement Learning \textbf{O}ptimization for \textbf{L}arge-scale \textbf{L}earning. \SysName{} caters to three primary user groups: tech pioneers aiming for cost-effective, fault-tolerant large-scale training, developers requiring flexible control over training workflows, and researchers seeking agile experimentation. \SysName{} is built upon the following key modules to effectively serve these user groups:
(1) A single-controller architecture combined with an appropriate abstraction of the \texttt{Parallel Worker} simplifies the development of the training pipeline.
(2) The \texttt{Parallel Strategy} and \texttt{Data Transfer} modules enable efficient and scalable training.
(3) The \texttt{Rollout Scheduler} offers fine-grained management of each sample’s lifecycle during the rollout generation stage.
(4) The \texttt{Environment Worker} and \texttt{Reward Worker} support rapid and flexible experimentation with agentic RL algorithms and reward designs, respectively.
(5) \texttt{AutoDeviceMapping} allows users to assign resources to different models across various stages flexibly. The in-house training of a Mixture-of-Experts (MoE) model with over 200B total parameters using \SysName{} successfully scales to thousands of GPUs for around two weeks without interruption, demonstrating its scalability and fault tolerance. We benchmark \SysName{} on a multi-domain task with verifiable rewards and three agentic RL tasks to validate its usability and effectiveness in handling a wide range of RL scenarios.







\end{abstract}

\begin{figure}[!h]
    \centering
    \includegraphics[width=0.9\textwidth]{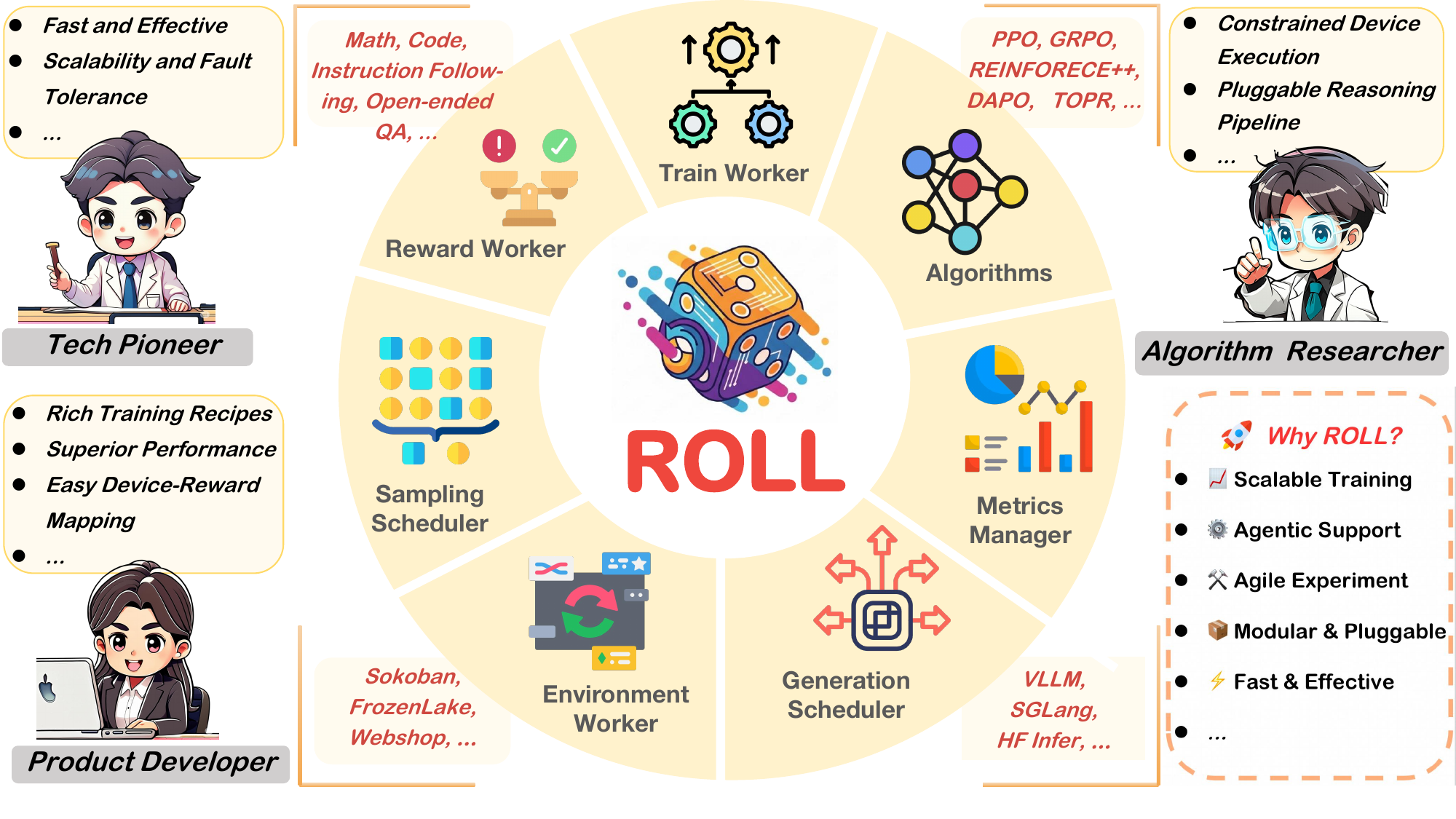}
    \caption{
    For three  primary user groups, we introduce an efficient, scalable, and user-friendly library \SysName{}, which  provides specific key features for large-scale RL optimization.}
    \label{fig:intro}
\end{figure}

\vfill

\newpage




%



\section{Introduction}
\label{sec:intro}

The successful adoption of Reinforcement Learning (RL) in Large Language Models (LLMs), pioneered by RL from Human Feedback, has instigated the development of advanced RL techniques for reference alignment~\citep{ouyang2022training,bai2022training}, reasoning enhancement~\citep{deepseekr1,qwq32}, and agentic tool use~\citep{skyrlv0}. Many leading LLMs including OpenAI o4~\citep{openaio4}, QwQ~\citep{qwq32}, Seed1.5-thinking~\citep{seed-thinking}, and Deepseek-R1~\citep{deepseekr1} have all leveraged RL to achieve outstanding performance across a range of AI tasks, including coding~\citep{openr1_codeforces_dataset}, mathematics~\citep{maxwell_jia_2024_aime_dataset}, and tool use~\citep{pan2024trainingsoftwareengineeringagents,verl-agent}.



Existing RL optimization algorithms for LLM contain several types of paradigms as follows: 
\begin{itemize}[leftmargin=10pt,topsep=1pt, itemsep=2pt, itemindent=8pt]
      \item RL from human Feedback~\citep{knox2012learning,knox2009interactively,maclin2005giving,judah2010reinforcement}.
      \item RL with verifiable rewards (RLVR)~\citep{zelikman2022star,lambert2024t,DAPO}.
      \item RL with multi-turn agentic interaction~\citep{skyrlv0,zhou2024archer,abdulhai2025lmrl}.
\end{itemize}
They typically require maintaining multiple LLMs and orchestrating a multi-stage training pipeline. A standard RL training workflow involves up to four distinct LLMs~\citep{bai2022training, ouyang2022training}: the \texttt{Actor}, \texttt{Critic}, \texttt{Ref}, and \texttt{Reward} models (illustrated in Section~\ref{sec:bg-rl-for-llm}). Each training iteration contains three stages. \textit{\textbf{Generation}}: The \texttt{Actor} generates responses based on a batch of input prompts. In agentic RL settings, the \texttt{Actor} may also interact with the environment over multiple turns. \textit{\textbf{Inference}}: The \texttt{Critic}, \texttt{Ref}, and \texttt{Reward} models perform forward passes over the generated responses to compute supervision signals or reward estimates. Recent RL endeavors have simplified it by reducing the number of LLMs in this stage even removing this stage~\citep{dpo,2ddpo,shao2024deepseekmath}. \textit{\textbf{Training}}: The \texttt{Actor} and \texttt{Critic} models update the parameters with the reward signal obtained in the inference stage. In certain RL algorithms~\citep{DAPO,shao2024deepseekmath,li2023remax}, the \texttt{Critic} model remains inactive. Most RL optimization approaches still fall within the broader family of these multi-model, multi-stage training paradigms. To support efficient RL optimization for LLMs, numerous system frameworks~\citep{Harper_NeMo_a_toolkit,hu2024openrlhf,zhong2024rlhfuseefficientrlhftraining,mei2024realhf,lei2024puzzle,StreamRL} have been proposed.
However, most of these efforts introduce several classical system design approaches including single-controller~\citep{sheng2024hybridflow}, colocation~\citep{mei2024realhf}, and disaggregated architectures~\citep{StreamRL} to accelerate RL training for LLMs.

Inspired by these seminal efforts, 
as shown in Figure~\ref{fig:intro},
we introduce \SysName{}, an efficient, scalable, and user-friendly library crafted to supercharge RL optimization for large-scale learning. \SysName{} delivers several key features to serve for three primary user groups. For \textbf{tech pioneers}, \SysName{} supports fast, cost-effective, scalable, and fault-tolerant RL training in large-scale GPU clusters with heterogeneous hardware. For \textbf{product developers}, \SysName{} provides flexible and fine-grained control to route input samples to the appropriate agent environments, reward workers, and devices, delivering strong performance with minimal engineering effort. For \textbf{algorithm researchers}, \SysName{} facilitates efficient training on resource-constrained GPU setups and enables agile experimentation with new ideas through well-designed abstractions of RL training pipelines.

Specifically,
\SysName{} consists of following pivotal key modules that empower its advanced features. 
\begin{itemize}[leftmargin=10pt,topsep=1pt, itemsep=2pt, itemindent=8pt]
\item We build upon the single-controller architecture proposed in~\citep{sheng2024hybridflow} and introduce a well-defined abstraction of the \texttt{Parallel Worker} to enable a flexible and modular RL training pipeline, thus easing the experimentation with new ideas.
\item We introduce the optimized \texttt{Parallel Strategy} and \texttt{Data Transfer} to enable execution on resource-constrained devices, as well as fast, scalable, and fault-tolerant training.
\item We provide the \texttt{Rollout Scheduler} to support fine-grained lifecycle management of each prompt sample during the generation stage, simplifying the orchestration of execution flow across response generation, environment interaction, and reward computation.


\item We dedicate the \texttt{Environment Worker} and \texttt{Reward Worker} to provide efficient and scalable agentic environment interaction and reward computation.

\item We implement the \texttt{Resource Pool} and leverage the \texttt{AutoDeviceMapping} to achieve efficient worker placement and optimized resource allocation. 
\end{itemize}


Our \SysName{} is built atop Ray~\citep{ray} and integrate existing LLM optimization systems including vLLM~\citep{vllm}, SGLang~\citep{sglang}, DeeepSpeed~\citep{Deepspeedautotuning}, and Megatron~\citep{shoeybi2019megatron}. Our in-house training of 200B+ MoE models on thousands of GPUs over two weeks without interruption demonstrates the efficiency and fault tolerance of \SysName{} in scalable RL training. Additionally, we benchmark \SysName{} on a multi-domain RLVR task encompassing code, math, and other verifiable domains, as well as on three agentic RL tasks, to validate its correctness and usability.

\section{Background}
\label{sec:background}
\subsection{RL for LLMs}
\label{sec:bg-rl-for-llm}
RL is a pivotal technique adopted in post-training for LLMs. Here, we brief several key concepts in RL for LLMs, followed by the workflow to training RL for LLMs. 

\noindent\textbf{Key Concepts.} 
RL training for LLMs typically employs policy gradient methods, particularly PPO (Proximal Policy Optimization) and its variants. The training pipeline usually consists of several key components: an \texttt{Actor} model that generates responses, a \texttt{Critic} that estimates value functions, a \texttt{Ref} model for preventing excessive divergence from initial behaviors, and a \texttt{Reward} model that evaluates response quality.
For a given prompt, the \texttt{Actor} model continuously generates a trajectory of tokens until the termination criteria yield the response. 
In this context, each token generated by the model represents an action in the RL framework, where the optimization objective is to maximize the expected cumulative reward by adjusting the policy (i.e., \texttt{Actor}) to generate sequences that better align with human preferences and task requirements.
The reference model (\texttt{Ref}) is usually initialized from the actor model and its weight is usually frozen during training. It serves as a regularization to ensure the actor model does not deviate overly from its initialized state. The reward model (\texttt{Reward}) is to provide a signal to guide the \texttt{Actor} to generate responses that align with specific goals, e.g., human preferences, tool use, and mathematical and code reasoning. It can be trained on human-labeled preference data using an LLM or we can use rule-based verification or sandbox execution to derive the reward value.
The critic model (\texttt{Critic}) estimates the value function in RL and evaluates the expected future rewards of the current state (i.e., the generated text sequence so far) to help reduce variance in policy gradient updates and guide the policy optimization of the \texttt{Actor}. 



\noindent\textbf{Optimization Workflow.} Each iteration in RL optimization for LLMs contains the generation, inference, and training stages as follows:

\textit{Generation Stage}: The actor model interacts with the environment and generates responses for a batch of prompts. This process involves prefill phase, decoding phase, and environment interaction phase. The prefill phase is a compute-bound GPU task, which proceeds the prompt to compute its key-value cache. The decoding phase is a memory-bound GPU task, autogressively generating tokens until meeting the termination criterion. The environment interaction phase involves executing complex environments and facilitating interactions between these environments and the actor model, utilizing intensive CPU resources. 


The single-turn tasks including mathematical and code reasoning, the actor model usually entails stateless environment interactions and only contains the prefill and decoding phase. In multi-turn tasks such as tool use, the actor model engages in multiple rounds of interaction with the environment, making the environment interaction phase a significant performance bottleneck.


In the generation stage, a rollout sample consists of tokens produced during prefill, decoding, and environment interaction and is utilized for later inference and training stages. A batch of responses are generated in this stage to accelerate convergence, albeit at the cost of substantial computational overhead.




\textit{Inference Stage}: 
During the RL training process, each generated sequence from the actor model is evaluated through a single forward pass by the reference, critic, and reward models. The reference model provides KL penalties to prevent excessive policy deviation, the critic model estimates value scores for advantage computation, and the reward model assigns quality scores. These outputs are then combined to compute the final training objective, which typically includes policy loss, value loss, and KL penalty terms.
The above process only entails the prefill phase, a compute-bound process. 

One exception case is the reward computation. The LLM-based reward computation can be considered as the prefill phase and runs on GPUs. The computation of the verifiable rewards including rule-based mathematical verification, and sandbox verification, is similar to the environment interaction phase and usually needs many CPU resources to obtain the reward targets quickly.

\textit{Training Stage}: The actor and critic model are updated with the produced samples in the generation stage and the reward signals in the inference stage. The updated parameters are synchronized for the generation stage in the subsequent iteration. Compared with the generation and inference stages, the training stage usually consumes substantial GPU memory and necessitates a variety of LLM parallelization strategies to enable its efficient execution.




\subsection{System Optimization for RL-enhanced LLMs} 
\noindent\textbf{Training.} LLM training can be accelerated by 5D parallelism including Data Parallelism (DP)~\citep{rajbhandari2020zero,zhao2023pytorchfsdp}, Tensor Parallelism (TP)~\citep{shoeybi2019megatron}, Pipeline Parallelism (PP)~\citep{huang2019gpipe}, Context Parallel (CP)~\citep{li2021sequenceparallelism}, and Expert Parallel (EP)~\citep{rajbhandari2022deepspeed}. Moreover, ZeRO~\citep{rajbhandari2020zero}, activation recomputation~\citep{chen2016recomp}, and offloading~\citep{zero-offload} can be adopted to alleviate the memory overhead. 

\noindent\textbf{Inference/Generation.} Many efficient LLM serving frameworks including SGLang~\citep{sglang} and vLLM~\citep{vllm} support DP, TP, PP, and EP efficiently. Besides, many LLM serving optimization works optimize the attention computation~\citep{li2023distflashattn,li2025mminference} and KV cache usage~\citep{KIVI,KVQuant,rethink-kv-sys}. 

\noindent\textbf{RL Optimizations.} RL training for LLMs possesses distinct computations, including generation, inference, and training and different sizes of LLMs. Particularly, the \texttt{Actor} model performs the generation and training stage, the \texttt{Critic} model performs the training and inference stage, the \texttt{Ref} model performs the inference stage, and the \texttt{Reward} model performs the inference stage. Thus, distinct parallel strategies can be tailored for different models across various stages to maximize the overall performance. NeMo~\citep{Harper_NeMo_a_toolkit} and OpenRLHF~\citep{hu2024openrlhf} divide the GPU cluster into several partitions and allocate them to different stages. In each stage, they run LLMs with optimized parallelization strategies. To improve the resource utilization, Verl~\citep{sheng2024hybridflow}, RLHFuse~\citep{zhong2024rlhfuseefficientrlhftraining}, ReaL~\citep{mei2024realhf}, and
PUZZLE~\citep{lei2024puzzle} colocate LLMs of different stages in the same resource pool. StreamRL~\citep{StreamRL} proposes to disaggregate the training and generation stages and asynchronously run the generation and training stages in a pipeline manner. Furthermore, the rollout generation can be accelerated owing to the high memory-bandwidth advantages in the inference cluster.


\subsection{RL Algorithms}
\paragraph{RL from Human Feedback.} 



The early success of RL optimization for LLMs is to guide LLMs in attaining human preferences. 
In the early stages, RL from Human Feedback (RLHF) methods mainly centered around learning directly from human rewards \citep{knox2012learning,knox2009interactively}, learning from action advice \citep{maclin2005giving}, or learning from action critique \citep{judah2010reinforcement}. For example, TAMER~\citep{warnell2018deep} interprets human feedback as samples of the optimal action-value function. The COACH~\citep{arumugam2019deep} considers the human feedback in a policy-dependent manner. Recently, after the release of ChatGPT, many RLHF methods~\citep{ouyang2022training,schulman2017proximal} have been proposed to align LLMs with human preferences and values, which typically includes three phases: supervised fine-tuning, reward model training, and policy optimization. However, these RLHF methods necessitate considerable human-annotated samples to train a reward model, preventing from their widespread adoption. 

\paragraph{RL with Verifiable Rewards.}
Some researchers~\citep{zelikman2022star,zhang2024openrft,lambert2024t,deepseekr1,DAPO} propose the RL with Verifiable Rewards (RLVR) on several representation reasoning tasks (e.g., math, code).
Specifically, the accuracies of these reasoning tasks are generally determined by whether the final answer is correct. This approach stems from the fact that reliably evaluating intermediate steps remains difficult, especially when those steps lack labeled ground truth. For example, researchers often employ rule-based policies to assess solutions in mathematical tasks, while they use the sandbox to judge whether the generated code successfully passes all test cases in coding tasks. 
In some cases, it is difficult to obtain the correctness of the answers, so the 
LLM-as-a-Judge~\citep{son2024llmasajudgerewardmodel} is adopted, which uses an LLM to identify the correctness of the generated answer. 
Recently, the dynamic sampling~\citep{DAPO} strategy has been widely used to filter samples based on the difficulties and improve the reasoning performance.

\paragraph{RL with Multi-turn Agentic Interaction.}

Unlike single-turn settings, where LLMs only perform one-pass response generation without ongoing environment interaction, multi-turn RL targets at more realistic agent scenarios~\citep{zhou2024archer,abdulhai2025lmrl}. Specifically, LLM-based agents need to perform a trajectory of actions to accomplish certain tasks, i.e., managing a  terminal~\citep{liu2024agentbench}, traversing web-based interfaces~\citep{zhou2024webarena}. The slow execution of environments, the difficulty in obtaining reward feedback from actions, and the complex interactions between environments and LLMs collectively pose a significant challenge to adopting RL optimization for LLMs in multi-turn agentic interaction scenarios.

\section{Key Features in \SysName{}}
\label{sec:key-feature}
We provide several key features to support efficient execution and user-friendly RL development. Hereafter, we discuss these key features from the dimension of the user groups. Particularly, we concentrate on the user experience of tech pioneers, product developers, and algorithm researchers. Moreover, we elaborate on the specifications for the agentic RL training pipeline.

\subsection{Tech Pioneer}
The tech pioneers seek the leading role in the LLM community and they possess a large-scale GPU cluster to facilitate the scalable RL training of LLMs. The advantages of \SysName{} manifest in three aspects to attract such a user group.

\begin{denseitemize}

\item \textbf{Fast and Cost Effective}: \SysName{} can fully exploit the high-performance hardware resources to expedite the RL training and achieve considerable training cost and time reduction in a large GPU cluster. 
\item \textbf{Scalability and Fault Tolerance}: \SysName{} supports a wide range of LLM training and serving optimization techniques, enabling the scalable training of a 200B-parameter model across thousands of GPUs without interruption for about two weeks. It also features an efficient checkpoint and resumption mechanism, allowing the training task to be restarted with minimal engineering effort. 

\item \textbf{Flexible Hardware Usage}: \SysName{} supports running RL training across a variety of hardware types. Users can choose between colocation or disaggregation, and configure synchronous or asynchronous execution modes, to fully leverage the advantages of different hardware architectures.
\end{denseitemize}

\subsection{Product Developer}
The product developers have enough GPUs to conduct RL training for in-house LLMs and they focus on configuring the task and reward to enhance LLMs with human alignment, reasoning capability, tool use, and business metrics. We recommend the product developers to choose \SysName{} for the following reasons.

\begin{denseitemize}
\item \textbf{Diverse and Extensible Rewards/Environments}: \SysName{} implements a set of \texttt{Reward Worker}s and \texttt{Environment Worker}s. Product developers can readily customize their own reward and environment by building upon our in-situ implementations.

\item \textbf{Compositional Sample-Reward Route}: \SysName{} provides a user-friendly interface to control the prompt sampling ratio across different tasks and dynamically route each sample to the corresponding \texttt{Reward Worker} (e.g., mathematical verifiers, sandbox environments, and LLM-as-a-judge). As production-level LLMs often encompass a diverse range of capabilities, this feature enables developers to optimize model performance across a mixture of domains and tasks.
\item \textbf{Easy Device-Reward Mapping}: \SysName{} develops the device-reward mapping interface to easily configure the device mapping of the Reward Worker. This feature isolates the reward computation from other computational workloads in multi-task RL training for LLMs, preventing interference and performance bottlenecks. 

\item \textbf{Rich Training Recipes}: \SysName{} provides various RL algorithms, LLMs, tasks, and datasets to reduce the engineering effort required for developing new training features.

\item \textbf{Superior Performance}: \SysName{} consists of a set of tuned training configurations that reach satisfactory performance across many tasks, relieving the burden of laborious hyperparameter search. 
\end{denseitemize}

\subsection{Algorithm Researcher}
Most algorithm researchers have access to a limited number of GPUs, and they require flexible, fine-grained control over each component of RL training for LLMs in order to efficiently experiment with new ideas. \SysName{} is well-suited for this purpose, offering the following key features.

\begin{denseitemize}
\item \textbf{Constrained Device Execution}: \SysName{} enables efficient training on constrained GPU resources via a set of memory optimization techniques, including single-GPU setups. This allows algorithm researchers to conduct multiple trial-and-error experiments and obtain timely feedback without requiring extensive high-grade GPUs.

\item \textbf{Pluggable Reasoning Pipeline}: \SysName{} abstracts each stage of the RL training pipeline at an appropriate level of granularity, enabling agile experimentation with new ideas. Researchers can flexibly orchestrate the execution of individual stages, facilitating the implementation and customization of diverse RL algorithms.
\item \textbf{Transparent Experimentation}: \SysName{} provides transparent logging and monitoring capabilities, making it easy to track and analyze each experiment.
\item \textbf{Fair Academic Baselines}: \SysName{} offers classical algorithms, models, and tasks to facilitate fair baseline comparisons on standard benchmarks.

\end{denseitemize}

\subsection{Specifications for Agentic RL}
The recent surge in agentic RL calls for efficient support for agent-based RL training with LLMs. To address this, we equip \SysName{} with the following features to enable scalable agentic RL training.

\begin{denseitemize}
\item \textbf{Scalable Multi-Turn Agent-Environment Interaction}: Inspired by RAGEN~\citep{ragen}, \SysName{} supports multi-turn interaction between agents and environments, scaling to long-horizon tasks.

\item \textbf{Sample-wise Scalable Environments}: \SysName{} flexibly performs environment scaling to match the size of the input sample to enable the high-throughput rollout. 

\item \textbf{Asynchronous Parallelized Agent-Environment Interaction}: \SysName{} performs environment execution and actor generation asynchronously through sample-wise environment management, and enables parallelized environment execution via environment scaling, reducing GPU idle time and maximizing resource utilization.

\end{denseitemize}



\section{Framework Design}
In this section, we discuss the design of \SysName{} to underpin relevant key features discussed in Section~\ref{sec:key-feature}. 

\begin{figure}[t]
    \begin{subfigure}{.46\textwidth}
        \centering
        \includegraphics[width=\textwidth]{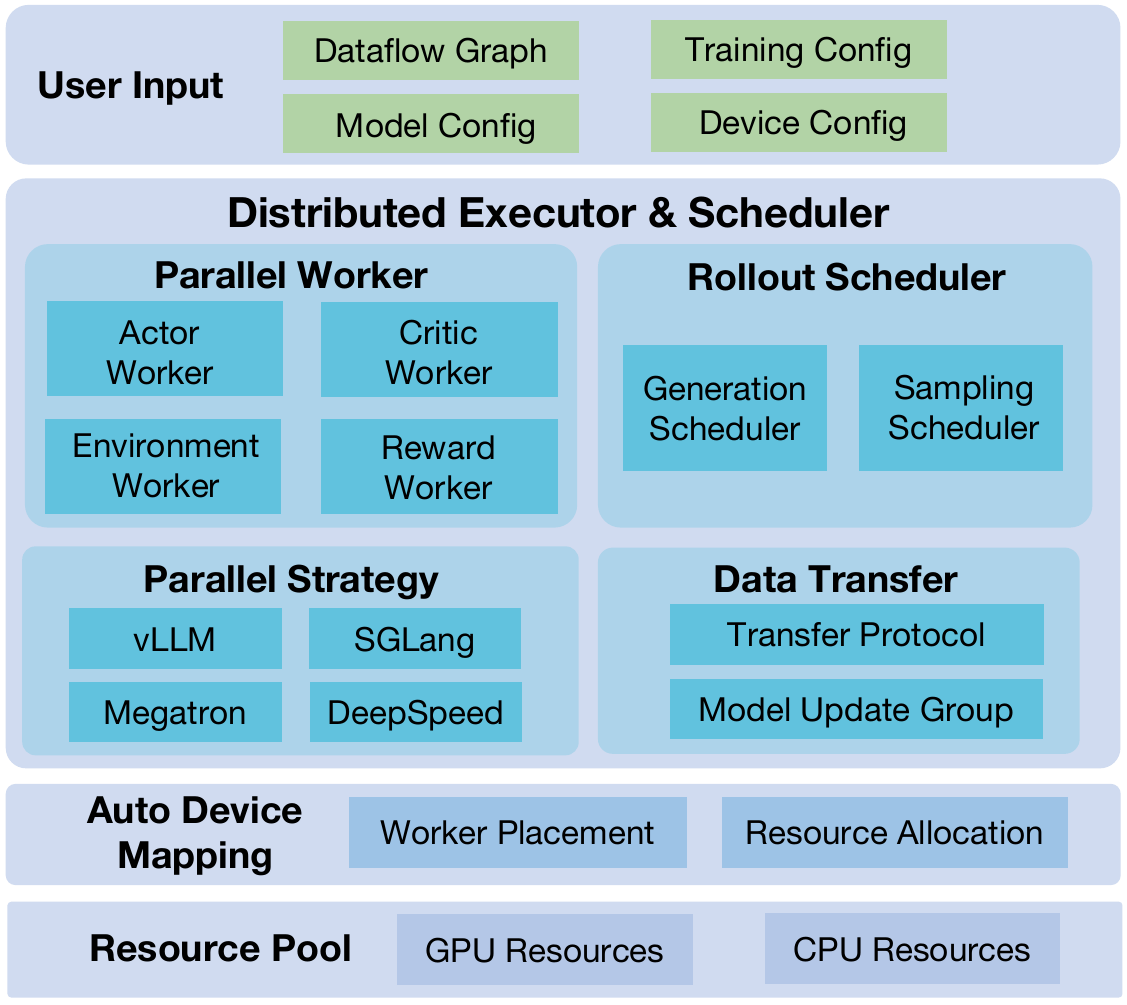}
        \caption{Architecture}
        \label{fig:arch}
    \end{subfigure}%
    \hfill
    \begin{subfigure}{.535\textwidth}
        \centering
        \includegraphics[width=\textwidth]{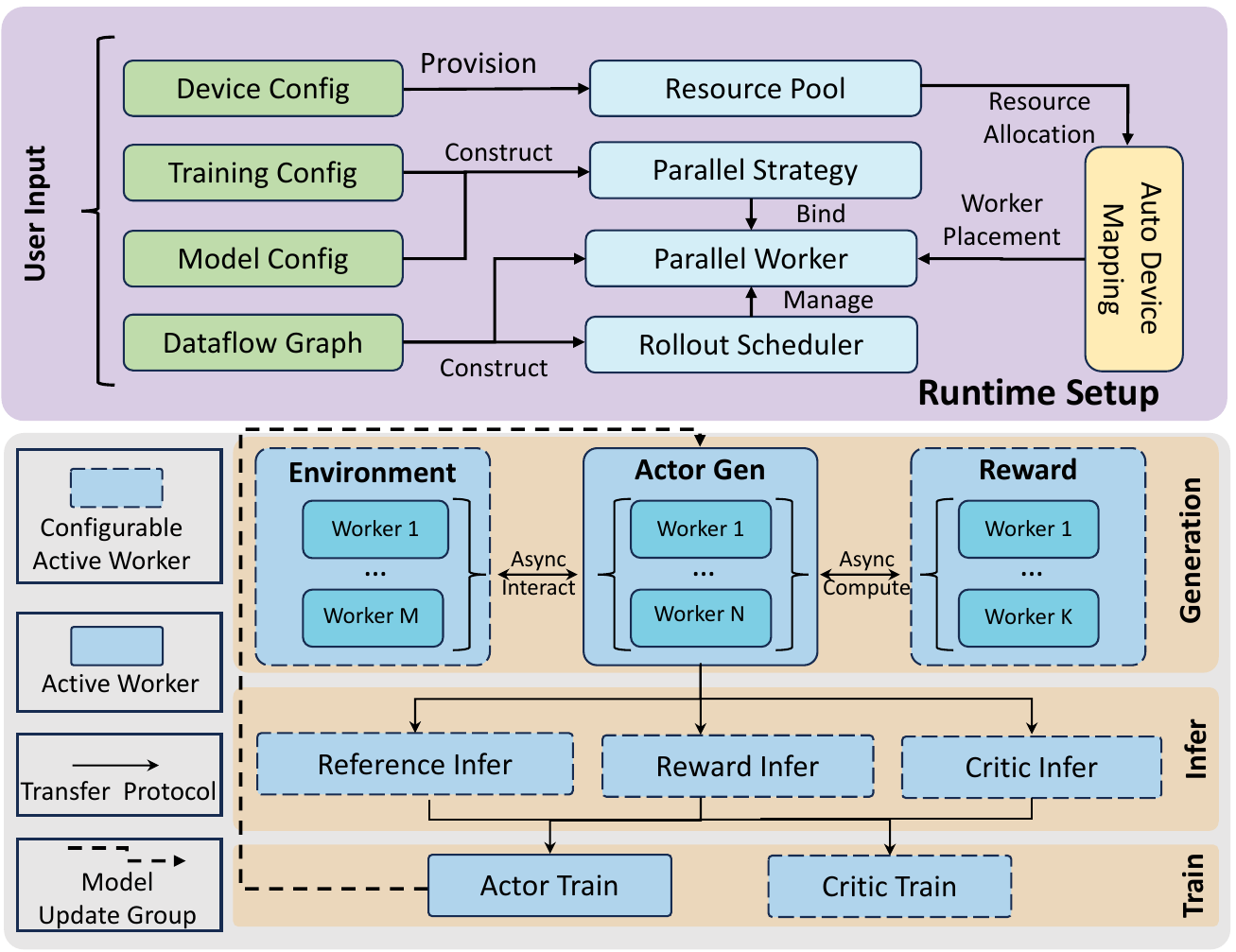}
        \caption{Workflow}
        \label{fig:workflow}
    \end{subfigure}
    \caption{(a) The architecture of \SysName{}, which consists of the user input layer, a distributed executor \& scheduler, an Auto Device Mapping module, and a resource pool. (b) The runtime setup and the training workflow of \SysName{}.}
    \vspace{-3mm}
    \label{fig:architecture_and_workflow}
\end{figure}

\subsection{System Architecture and Modules}
\paragraph{Architecture.} Figure~\ref{fig:arch} illustrates the architecture of \SysName{}. \SysName{} takes as input a user-defined RL dataflow graph along with its associated configurations. Based on this input, the distributed executor and scheduler orchestrate the workers and schedulers. The \texttt{AutoDeviceMapping} module manages resources within the provisioned resource pool and efficiently binds workers and schedulers to its allocated resources.

\paragraph{Parallel Worker.} \texttt{Parallel Worker} is the owner of a collective of resources (i.e., \texttt{PlacementGroup} in ray), and \SysName{} uses the \texttt{Cluster} to represent a group of \texttt{Parallel Worker}s that share the same role (e.g., actor training, critic inference) in the RL training to simplify the collective management of these workers. \SysName{} provides several types of \texttt{Parallel Worker}s. The \texttt{Actor Worker} can be instantiated to serve as either an \texttt{Actor} or a \texttt{Ref}. The \texttt{Critic Worker} implements the functionality of the \texttt{Critic}, while the \texttt{Reward Worker} handles the \texttt{Reward} component, delivering various reward computation methods including rule-based verification~\citep{he2025deepmath}, sandbox execution~\citep{dou2024multiprogramminglanguagesandboxllms}, and LLM-as-a-Judge~\citep{son2024llmasajudgerewardmodel}. The \texttt{Environment Worker} supports multi-turn interaction between various types of environments and LLMs.


\paragraph{Parallel Strategy.} RL training in \SysName{} encompasses the training, inference, and generation stages. We integrate MegatronCore and DeepSpeed to accelerate LLM training, supporting advanced 5D parallelism strategies including DP, PP, TP, CP, and EP. \SysName{} also supports ZeRO2, ZeRO3, and ZeRO-offload~\citep{zero-offload} thanks to DeepSpeed~\citep{rajbhandari2022deepspeed}. Additionally, we provide gradient checkpointing and offloading strategies to significantly reduce GPU memory consumption, enabling efficient execution on resource-constrained devices. For the inference and generation stage, we integrate the vLLM~\citep{vllm} and SGLang~\citep{sglang} to equip \SysName{} with TP, EP, PP to expedite the inference and generation stage. 

\paragraph{Rollout Scheduler.}The \texttt{Rollout Scheduler} allows users to schedule the lifecycle of each request at the granularity of individual samples, rather than batches, during the generation stage. Particularly, the \texttt{Rollout Scheduler} can dynamically add and abort requests based on current resource availability and the progress of response generation.

\paragraph{Data Transfer.}The \texttt{Transfer Protocol} was first introduced in HybridFlow~\citep{sheng2024hybridflow}, and we reuse it to reshard the input and output data across different stages. The \texttt{ModelUpdateGroup} is implemented to enable fast parameter synchronization between the training and generation/inference stages, supported by the NCCL communication backend, even in collocated training scenarios.

\paragraph{AutoDeviceMapping and Resource Pool.} The \texttt{AutoDeviceMapping} module orchestrates a set of CPU and GPU resources in the \texttt{Resource Pool} and binds them to the workers and scheduler. 


\subsection{System Workflow}
Figure~\ref{fig:workflow} depicts the workflow including the runtime setup (\textbf{Top}) and the training iteration (\textbf{Bottom}). 

\noindent\textbf{Runtime Setup.} \SysName{} provisions a resource pool comprising GPU and CPU resources based on the provided device configurations. Guided by the RL dataflow, it creates a \texttt{Rollout Scheduler} and multiple \texttt{Parallel Worker}s. The \texttt{Rollout Scheduler} oversees the lifecycle of each prompt sample request during the generation stage. Based on the training and model configurations, \SysName{} instantiates the \texttt{Parallel Strategy} to decide the parallelization strategy and execution backend for each \texttt{Parallel Worker}. Once the \texttt{Parallel Worker}s are established, \SysName{} follows the device mapping configurations specified by the users and employ \texttt{AutoDeviceMapping} to allocate resources from the resource pool to the respective \texttt{Parallel Worker}s.

\noindent\textbf{Training Iteration.} In the \textbf{generation} stage, a batch of samples is first fed to the \texttt{Rollout Scheduler} to generate responses. During this phase, the \texttt{Actor} model may interact with the \texttt{EnvironmentWorker} to perform multi-turn environment interactions in agentic RL tasks. It also invokes the \texttt{Reward Worker} to compute reward signals, enabling advanced sampling techniques (e.g., dynamic sampling~\citep{DAPO}) to enhance sampling efficiency. 

The subsequent \textbf{inference} stage involves forward passes by the \texttt{Critic}, \texttt{Reward}, and \texttt{Ref} models, provided they are activated in the RL dataflow graph. The \texttt{Transfer Protocol} then shards the responses from the generation stage and feeds them to each active \texttt{Parallel Worker}.

In the \textbf{training} stage, the \texttt{Critic} and \texttt{Actor} models update their parameters using the prepared reward signals. Besides, the \texttt{Actor} model also synchronizes model parameters with the generation stage via the \texttt{ModelUpdateGroup} in the next training iteration.



\subsection{How to Underpin Key Features}
We explain how our system modules in \SysName{} to support key features discussed in Section~\ref{sec:key-feature}. 
\paragraph{Single-Controller Pipeline.} We follow the hybrid programming model of HybridFlow~\citep{sheng2024hybridflow} to implement the training pipelines for RLHF, RLVR, and agentic RL within a single controller, simplifying the development and management of RL training workflows.

\paragraph{Worker Abstraction for RL Pipeline.} The abstractions of \texttt{Parallel Worker} and \texttt{RolloutScheduler} enable users to define and experiment with new pipelines with minimal engineering effort, by following our provided training workflow example. Particularly, the \texttt{Actor Worker}, \texttt{Critic Worker}, \texttt{Reward Worker}, and \texttt{Environment Worker} encapsulate the distinct roles in RL training. The well-defined abstraction allows users to concentrate on developing and customizing individual components without overhauling the entire codebase.

\paragraph{Optimized LLM Execution.} We fully capitalize on the advanced features of existing LLM execution engines, including DeepSpeed, Megatron, vLLM, and SGLang, to facilitate RL optimization in both large-scale GPU clusters and resource-constrained device environments.

\paragraph{User-defined Device Mapping.}
Prior RL systems including OpenRLHF~\citep{hu2024openrlhf} and NeMo~\citep{akter2025nemotron} enforce exclusive resource usage across different training stages. Recent research efforts~\citep{sheng2024hybridflow,zhong2024rlhfuseefficientrlhftraining} support the colocation of LLMs from different stages within the same device group. In \SysName{}, the \texttt{AutoDeviceMapping} module supports flexible, user-defined device mapping, allowing a single device to be shared across multiple LLMs from different stages. This enables users to reallocate a portion of the GPUs assigned to the generation stage of the \texttt{Actor} model to its training stage, improving overall resource utilization.

This capability stems from two key functionalities. First, \SysName{} is implemented on top of Ray, which allows us to bind each device to specific workers while also allowing multiple workers to share the same device. Second, the \texttt{ModelUpdateGroup} facilitates model synchronization across different stages. As previously discussed, a group of \texttt{Parallel Worker}s that share the same LLM role in RL training can be organized into a \texttt{Cluster}. Once synchronizing model parameters between the \texttt{Actor\_Train Cluster} and the \texttt{Actor\_Infer Cluster}, each worker in the training stage broadcasts its model parameters to corresponding workers in the generation stage in bucketed chunks, thereby improving the transfer speed. This design avoids enforcing co-location of training and inference processes, thereby supporting much more flexible, user-defined device mapping than prior RL systems~\citep{akter2025nemotron,hu2024openrlhf,sheng2024hybridflow,zhong2024rlhfuseefficientrlhftraining}

\paragraph{Sample-level Rollout Lifecycle Control.}
Most RL systems~\citep{akter2025nemotron,hu2024openrlhf,sheng2024hybridflow,zhong2024rlhfuseefficientrlhftraining} process a batch of prompt samples during the generation stage to improve the throughput. However, long-tail issues in the generation stage~\citep{zhong2024rlhfuseefficientrlhftraining} lead to imbalanced resource utilization across different workers. To address this, the \texttt{Rollout Scheduler} provides a request rollout lifecycle control in the granularity of each prompt sample during the generation stage.

The optimization of dynamical sampling provided by \SysName{} is a successful adoption of sample-level rollout lifecycle control. Dynamic sampling refers to the strategy of oversampling prompts and filtering out those with accuracy scores of either 1 or 0, while retaining only those that contribute effective gradients. Sample-level rollout lifecycle control can significantly accelerate dynamic sampling in three key aspects. (1) \texttt{Async Reward Computation}: \SysName{} removes the synchronization barrier between the generation and reward computation phases by initiating reward computation for completed samples immediately, rather than waiting for all prompts in the batch to finish the response generation. (2) \texttt{Add request}: \SysName{} continuously monitors worker completion states and dynamically dispatches new prompt samples based on real-time demand, thereby improving resource utilization. (3) \texttt{Abort Request}: Once the number of prompts yielding effective gradients reaches the target threshold, \SysName{} can proactively terminate other ongoing response generation tasks, reducing unnecessary generation overhead.



\paragraph{Sample-Wise Management of Rewards and Environments.}
The generation phase of the training workflow in Figure~\ref{fig:workflow} depicts the asynchronous reward computation and asynchronous environment interaction. \SysName{} can spawn multiple \texttt{Reward Worker}s and \texttt{Environment Worker}s based on job load at scale, distributing them across resource pools to prevent performance bottlenecks. Sample-level rollout lifecycle control allows users to flexibly route each sample to the corresponding \texttt{Reward Worker} and \texttt{Environment Worker}. 

\SysName{} leverages ray to support asynchronous reward computation. During RL training, multiple types of \texttt{Reward Worker}s including rule-based verification, sandbox execution, and LLM-as-a-Judge can be activated. These workers dynamically perform reward computation at runtime based on the current job load, while sample-level rollout control allows for flexible and compositional routing of samples to the appropriate \texttt{Reward Worker} on demand. Owing to \texttt{AutoDeviceMapping}, each \texttt{Reward Worker} is assigned to user-specified devices, simplifying the allocation of reward modules to hardware resources.

Similar to the \texttt{Reward Worker}, \SysName{} allocates sufficient resources to deploy scalable \texttt{Environment Worker}s and facilitates efficient interactions between \texttt{Actor} models and environments at scale. Furthermore, it supports parallelized environment interactions, enhancing environment throughput and reducing delays caused by waiting for responses. Sample-level rollout lifecycle control allows the \texttt{Actor} to process other samples without waiting for responses from the \texttt{Environment Worker}s. In this scenario, \SysName{} can asynchronously initiate new prompt samples for generation, thereby preventing resource underutilization. This mechanism is referred to as asynchronous environment interaction. Given that these \texttt{Environment Worker}s may be CPU-intensive, \SysName{} carefully distributes them across available resource pools to minimize interference with other workloads and among workers themselves.

\section{Experiments}

\subsection{RLVR Pipeline}

\paragraph{Data Collection.}
The experimental data for \SysName{} on RLVR pipeline is systematically curated from established sources across three domains: (1) Math domain: DeepMath-103K~\citep{he2025deepmath}, from which we sample 5,000 examples proportionally according to the difficulty; (2) Code domain: KodCode~\citep{xu2025kodcode}, from which we first filter out low-quality data and evenly sample 2,000 records based on the difficulty; (3) General domain: Multi-subject-RLVR~\citep{su2025crossing}, Nemotron-CrossThink~\citep{akter2025nemotron} and RLVR-IFeval~\citep{lambert2024t}, from which we intentionally remove low-quality data.

\paragraph{Training Setting.}
We experiment with two LLMs: Qwen2.5-7B-base and Qwen3-30B-A3B-base. For policy optimization, we utilize PPO loss where the advantage value is computed using REINFORCE returns instead of GAE-based estimates~\citep{schulman2015high}. The sampling ratio across domains is set to 40\% for math, 30\% for code, and 30\% for general reasoning. We incorporate rule-based verification, sandbox execution for code, and both rule-based verification and LLM as a judge for general reasoning. More detailed training configurations can be found in the following files\footnote{\url{https://github.com/alibaba/ROLL/blob/main/examples/qwen2.5-7B-rlvr_megatron/rlvr_config.yaml}}$^{,}$\footnote{\url{https://github.com/alibaba/ROLL/blob/main/examples/qwen3-30BA3B-rlvr_megatron/rlvr_config.yaml}}.



 \paragraph{Performance.}

 \begin{figure}[t]
    \centering
    \includegraphics[width=1.0\textwidth]{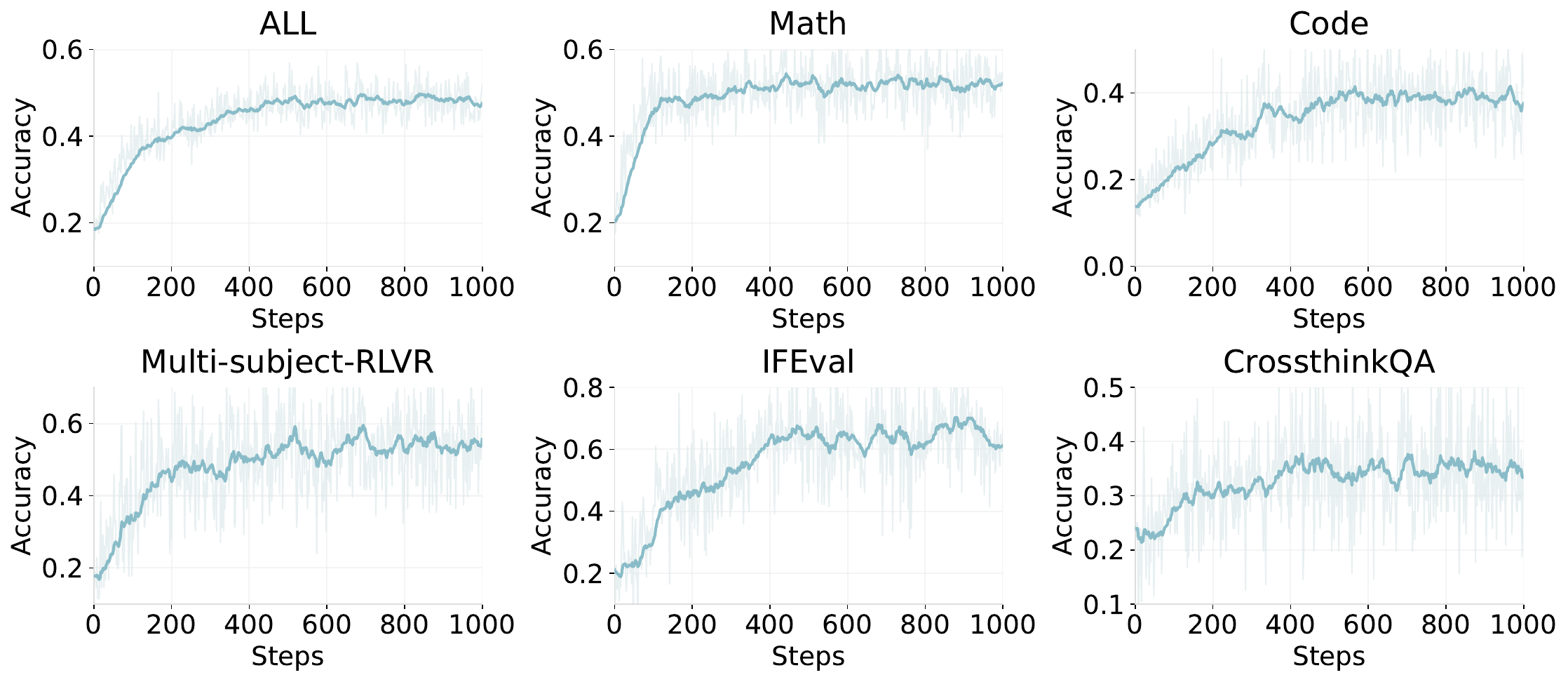}
    \caption{Accuracy Trends Across Different Tasks on Qwen2.5-7B-Base.}
    \vspace{-3mm}
    \label{fig: qwen2.5_correct}
\end{figure}

 \begin{figure}[t]
    \centering
    \includegraphics[width=1.0\textwidth]{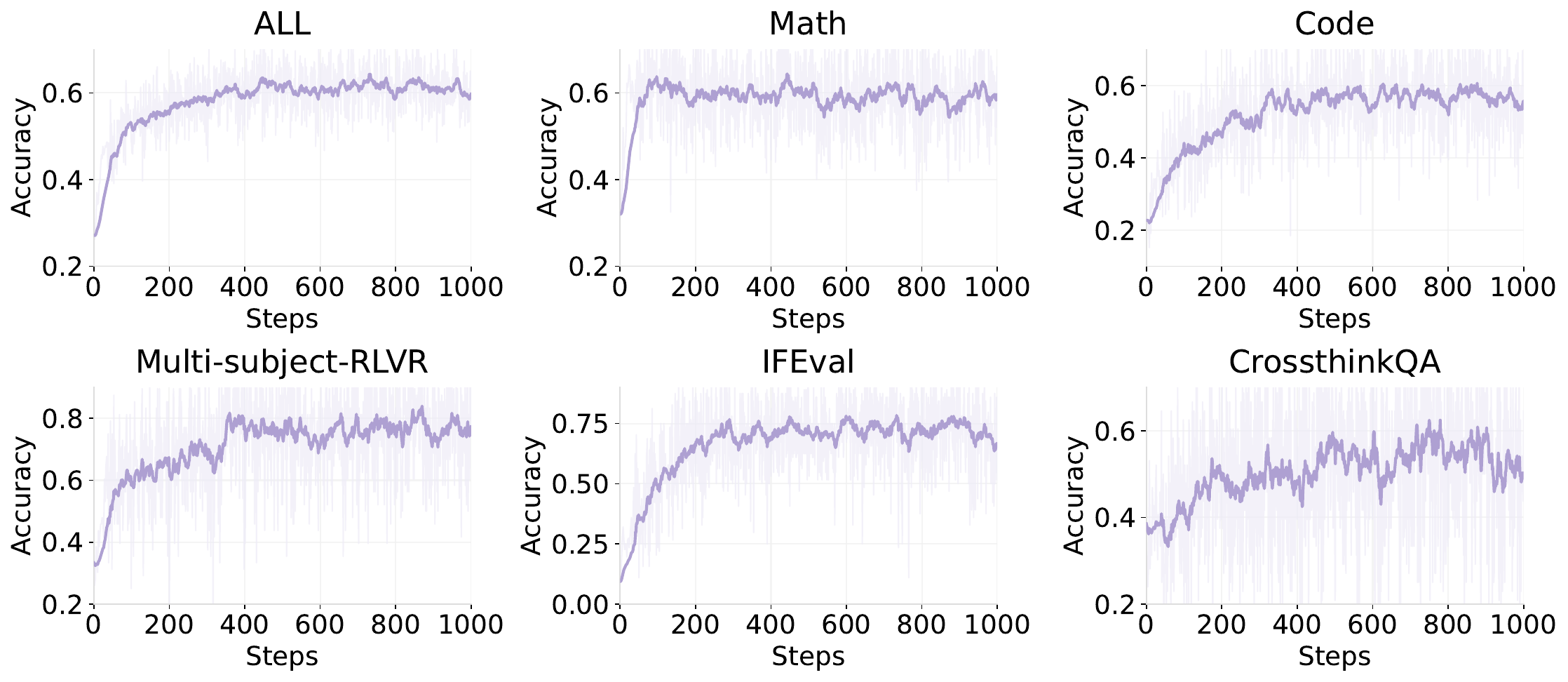}
    \caption{Accuracy Trends Across Different Tasks on Qwen3-30B-A3B-Base.}
    \label{fig: qwen3_correct}
\end{figure}

As shown in Figure~\ref{fig: qwen2.5_correct}, the accuracy of the Qwen2.5-7B-Base model on average rises from 0.18 to 0.52, representing a 2.89$\times$ improvement. Task-level analysis reveals marked gains in math reasoning (from 0.20 to 0.53) and code generation (from 0.13 to 0.41), highlighting the correctness and effectiveness of \SysName{} in task-specific tasks.


Figure~\ref{fig: qwen3_correct} illustrates the accuracy of Qwen3-30B-A3B-Base on different tasks, which improves from 0.27 to 0.62, yielding a 2.30 \(\times\) increase. Although the Qwen3-30B-A3B-Base model, which employs a mixture-of-experts architecture, exhibits greater accuracy fluctuations during training compared to the Qwen2.5-7B-Base model, it still demonstrates a clear upward trend and ultimately achieves superior performance. Overall, both models exhibit stable and consistent accuracy improvements throughout the training process without experiencing model collapse, indicating the robustness and practicality of \SysName{}.


\subsection{Agentic Pipeline}

We conduct extensive experiments across three distinct environments to rigorously evaluate the capabilities and adaptability of our agentic pipeline.

\subsubsection{Sokoban}

 \begin{figure}[t]
    \centering
    \includegraphics[width=\textwidth]{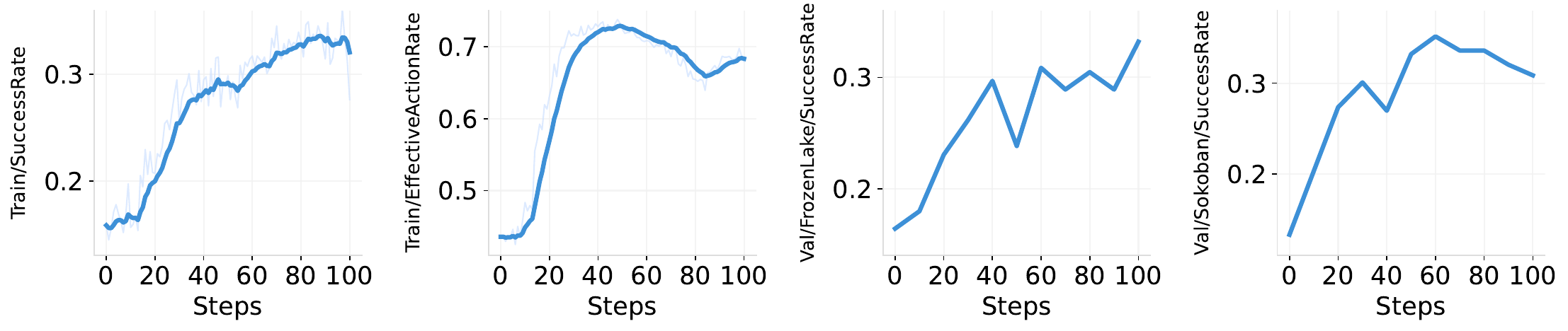}
    \caption{
        Performance metrics for the SimpleSokoban environment training. 
        \textit{SuccessRate} denotes the success rate of reaching the goal. 
        \textit{EffectiveActionRate} represents the proportion of valid actions executed. 
    }
    \vspace{-3mm}
    \label{fig: SimpleSokoban}
\end{figure}

\paragraph{Environment Configuration.}
The \textit{Sokoban} environment is a classic puzzle where the agent pushes boxes onto target locations within a grid. We configure three variants: (1) \textit{SimpleSokoban}, a 6×6 grid with one box; (2) \textit{LargerSokoban}, an 8×8 grid with two boxes; and (3) \textit{SokobanDifferentGridVocab}, a 6×6 grid using different symbols. Actions allowed are directional moves (Up, Down, Left, Right).

\paragraph{Training Setting.}
We employ the Qwen2.5-0.5B-Instruct model as the base model for training in the \textit{Sokoban} environment. Training tasks are distributed across 8 GPUs, using a rollout batch size of 1024. For policy optimization, we utilize PPO loss where the advantage value is computed using REINFORCE returns instead of GAE-based estimates, incorporating advantage clipping at 10.0 and reward clipping at 20 to maintain training stability. A format penalty with a weight of -0.001 is applied to encourage properly formatted action outputs. More detailed training configurations can be found here\footnote{\url{https://github.com/alibaba/ROLL/blob/main/examples/qwen2.5-0.5B-agentic_ds/agentic_val_sokoban.yaml}}.

\paragraph{Performance.}
Figure~\ref{fig: SimpleSokoban} presents the training results in the \textit{SimpleSokoban} environment. The model achieves a substantial performance gain, with the success ratio in training increasing from 16.8\% to 26.0\%. The success rate in the validation environment rises from 13.3\% to 35.2\%, and the proportion of effective actions grows from 43.6\% to 73.4\%, indicating steady improvement in agent capabilities. Moreover, these gains generalize well to the \textit{FrozenLake} environment, demonstrating the robustness of our RL training framework.

\subsubsection{FrozenLake}
 \begin{figure}[t]
    \centering
    \includegraphics[width=\textwidth]{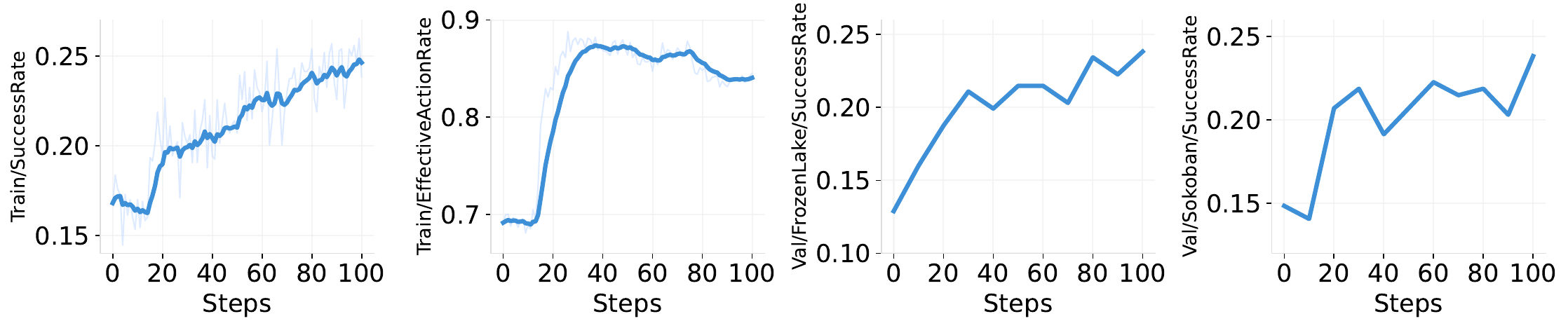}
    \caption{Performance metrics for the FrozenLake environment training.}
    \label{fig: FrozenLake}
\end{figure}

\paragraph{Environment Configuration.}
The FrozenLake environment requires an agent to navigate from a start to a goal position on a frozen surface while avoiding holes. The optional slippery ice mechanism introduces stochasticity by causing unintended movements, thus challenging the agent's adaptability to uncertainty.

\paragraph{Training Setting.}
We utilize the same Qwen2.5-0.5B-Instruct model and training configurations as the one in Sokoban environment for consistency. We refer readers to the repository for detailed training configurations\footnote{\url{https://github.com/alibaba/ROLL/blob/main/examples/qwen2.5-0.5B-agentic_ds/agent_val_frozen_lake.yaml}}.


\paragraph{Performance.}
Figure~\ref{fig: FrozenLake} presents the training results in the FrozenLake environment. The model demonstrates steady performance gains, with the success ratio in training increasing from 16.8\% to a peak of 26.0\%, representing a 55\% improvement. Concurrently, the proportion of effective actions rises from 69.1\% to a peak of 88.8\%, indicating enhanced action quality during training. On the validation set, the success rate demonstrates a corresponding pattern, rising from 12.9\% at the beginning of training to a maximum of 23.8\%. Meanwhile, the model also exhibits cross-environment transfer learning capabilities, with SimpleSokoban validation success rates reaching 23.8\% despite being trained exclusively on FrozenLake.

\subsubsection{WebShop}

 \begin{figure}[t]
    \centering
    \includegraphics[width=\textwidth]{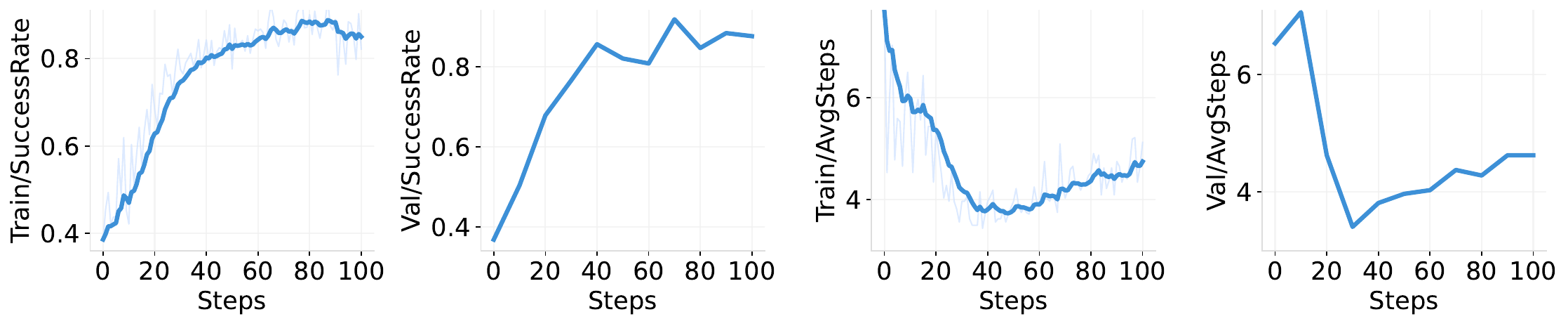}
    \caption{
    Performance metrics for the WebShop environment training. 
    \textit{AvgSteps} indicates the average number of steps required to complete the task, where fewer steps imply higher action efficiency.
    }
    \label{fig: webshop}
\end{figure}

\paragraph{Environment Configuration.}
The \textit{WebShop} environment simulates an online shopping task where the agent finds specific products using natural-language instructions. The agent performs iterative actions, including keyword searches, selecting product links, examining product details (e.g., description, features, size, color), and making purchase decisions. Actions vary by webpage context, and each trajectory is limited to 50 steps, highlighting the complexity of decision-making and instruction-following capabilities required.

\paragraph{Training Setting.}
We use the Qwen-2.5-7B-Instruct model for training in the \textit{WebShop} environment to support long interactions and rich context. The sequence length is set as 8192 tokens. We retain the REINFORCE algorithm and use the same clipping parameters for advantage estimation. We increase the format penalty to -0.05 to encourage well-formed responses. More detailed training configurations can be found here\footnote{\url{https://github.com/alibaba/ROLL/blob/main/examples/qwen2.5-0.5B-agentic_ds/agentic_val_webshop.yaml}}.


\paragraph{Performance.}
Figure~\ref{fig: webshop} shows a substantial improvement in task success rate, increasing from 37\% to over 85\% on both training and validation environments. The average number of actions per episode decreases from over 7 to around 4, indicating that the LLM learns to complete tasks more efficiently. Overall, LLMs can effectively possess the capability of task competence and operational efficiency to cope with real-world environments.

\section{Conclusion}
In this report, we introduce \SysName{}, a framework designed to optimize RL training for LLMs at scale. \SysName{} caters to three primary user groups: tech pioneers, product developers, and RL researchers. At its core, \SysName{} is built upon a suite of system modules, including \texttt{Parallel Worker}, \texttt{Rollout Scheduler}, \texttt{Parallel Strategy}, and \texttt{AutoDeviceMapping}, which together form the foundation of \SysName{}. Our extensive empirical evaluation demonstrates the effectiveness of \SysName{} in accelerating and scaling RL training for LLMs.

\clearpage
\section{Authors}

Within each role, authors are listed alphabetically.

\begin{multicols}{2}

\textbf{\textcolor{orange!80!black}{Project Lead}}\\
\vspace{-3mm}
\begin{itemize}[leftmargin=1.5em, itemsep=1pt]
    \item Weixun Wang
    \item Shaopan Xiong
\end{itemize}

\vspace{1em}
\textbf{\textcolor{orange!80!black}{Core Contributors}}\\
\begin{itemize}[leftmargin=1.5em, itemsep=1pt]
    \item Gengru Chen
    \item Wei Gao
    \item Sheng Guo
    \item Yancheng He
    \item Ju Huang
    \item Jiaheng Liu
    \item Zhendong Li
    \item Xiaoyang Li
    \item Zichen Liu
    \item Haizhou Zhao
\end{itemize}

\vspace{1em}
\textbf{\textcolor{orange!80!black}{Contributors}}\\
\begin{itemize}[leftmargin=1.5em, itemsep=1pt]
    \item Dakai An
    \item Lunxi Cao
    \item Qiyang Cao
    \item Wanxi Deng
    \item Feilei Du
    \item Yiliang Gu
    \item Jiahe Li
    \item Xiang Li
    \item Mingjie Liu
    \item Yijia Luo
    \item Zihe Liu
    \item Yadao Wang
    \item Pei Wang
    \item Tianyuan Wu
    \item Yanan Wu
    \item Yuheng Zhao
    \item Shuaibing Zhao
    \item Jin Yang
    \item Siran Yang
    \item Yingshui Tan
    \item Huimin Yi
    \item Yuchi Xu
    \item Yujin Yuan
    \item Xingyao Zhang
\end{itemize}

\vspace{1em}
\textbf{\textcolor{orange!80!black}{Supervision}}\\
\begin{itemize}[leftmargin=1.5em, itemsep=1pt]
    \item Lin Qu
    \item Wenbo Su
    \item Wei Wang
    \item Jiamang Wang
    \item Bo Zheng
\end{itemize}

\end{multicols}


\clearpage

\bibliographystyle{colm2024_conference}
\bibliography{colm2024_conference} 

\end{document}